\documentclass[sn-mathphys,Numbered,iicol]{sn-jnl}% Math and Physical Sciences Reference Style
\usepackage{graphicx}%
\usepackage{subcaption}
\usepackage{multirow}%
\usepackage{amsmath,amssymb,amsfonts}%
\usepackage{amsthm}%
\usepackage{mathrsfs}%
\usepackage[title]{appendix}%
\usepackage{xcolor}%
\usepackage{textcomp}%
\usepackage{manyfoot}%
\usepackage{booktabs}%
\usepackage{algorithm}%
\usepackage{algorithmicx}%
\usepackage{algpseudocode}%
\usepackage{listings}%
\usepackage{xcolor}
\definecolor{Dandelion}{RGB}{240, 183, 70}
\newcommand{\mysection}[1]{\vspace{2pt}\noindent\textbf{#1}}
\usepackage{xcolor, color, colortbl}         % colors
\definecolor{Highlight}{HTML}{39b54a}  % green

\usepackage{amssymb}% http://ctan.org/pkg/amssymb
\usepackage{pifont}% http://ctan.org/pkg/pifont
\usepackage{algorithm}
\usepackage{listings}
\usepackage{etoolbox}
\makeatletter
\AfterEndEnvironment{algorithm}{\let\@algcomment\relax}
\AtEndEnvironment{algorithm}{\kern2pt\hrule\relax\vskip3pt\@algcomment}
\let\@algcomment\relax
\newcommand\algcomment[1]{\def\@algcomment{\footnotesize#1}}
\renewcommand\fs@ruled{\def\@fs@cfont{\bfseries}\let\@fs@capt\floatc@ruled
  \def\@fs@pre{\hrule height.8pt depth0pt \kern2pt}%
  \def\@fs@post{}%
  \def\@fs@mid{\kern2pt\hrule\kern2pt}%
  \let\@fs@iftopcapt\iftrue}
\makeatother
\newcommand{\cmmnt}[1]{}

\usepackage[normalem]{ulem}

\usepackage{listings}
\usepackage{color}
\definecolor{codegreen}{rgb}{0,0.6,0}
\definecolor{codegray}{rgb}{0.5,0.5,0.5}
\definecolor{codepurple}{rgb}{0.58,0,0.82}
\definecolor{backcolour}{rgb}{1,1,1}
 
\lstdefinestyle{mystyle}{
    backgroundcolor=\color{backcolour},   
    commentstyle=\color{codegreen},
    keywordstyle=\color{magenta},
    numberstyle=\tiny\color{codegray},
    stringstyle=\color{codepurple},
    basicstyle=\footnotesize,
    breakatwhitespace=false,         
    breaklines=true,                 
    captionpos=b,                    
    keepspaces=true,                 
    numbers=left,                    
    numbersep=5pt,                  
    showspaces=false,                
    showstringspaces=false,
    showtabs=false,                  
    tabsize=2
}
 
\makeatletter

\newcommand{\Rmnum}[1]{\expandafter\@slowromancap\romannumeral #1@}
\makeatother

\usepackage{multirow}
\usepackage[symbol]{footmisc}
\usepackage{xspace}
\newcommand{\etal}{\emph{et al.}\xspace}
\newcommand{\eg}{\emph{e.g.}\xspace}
\newcommand{\ie}{\emph{i.e.}\xspace}
\usepackage{balance}
\usepackage{adjustbox}

\begin{document}

\title[Article Title]{SoccerNet 2023 Challenges Results}

\author*[1,2]{\fnm{Anthony} \sur{Cioppa$^{\dag}$}}\email{anthony.cioppa@uliege.be}
\author*[2]{\fnm{Silvio} \sur{Giancola$^{\dag}$}}\email{silvio.giancola@kaust.edu.sa}
\author[3,4,5]{\fnm{Vladimir} \sur{Somers$^{\dag}$}}
\author[1,6]{\fnm{Floriane} \sur{Magera$^{\dag}$}}
\author[7]{\fnm{Xin} \sur{Zhou$^{\dag}$}}
\author[8]{\fnm{Hassan} \sur{Mkhallati$^{\dag}$}}
\author[1]{\fnm{Adrien} \sur{Deliège$^{\dag}$}}
\author[1]{\fnm{Jan} \sur{Held$^{\dag}$}}
\author[2]{\fnm{Carlos} \sur{Hinojosa$^{\dag}$}}
\author[9]{\fnm{Amir M.} \sur{Mansourian$^{\dag}$}}
% ...
\affil[1]{\orgname{University of Liege (ULiège)}}
\affil[2]{\orgname{King Abdullah University of Science and Technology (KAUST)}}
\affil[3]{\orgname{Sportradar}}
\affil[4]{\orgname{UCLouvain}}
\affil[5]{\orgname{EPFL}}
\affil[6]{\orgname{EVS Broadcast Equipment}}
\affil[7]{\orgname{Baidu Research}}
\affil[8]{\orgname{Université Libre de Bruxelles (ULB)}}
\affil[9]{\orgname{Sharif University of Technology}}
% ...
% Number of authors:  102
% Number of institutions:  40


\abstract{
The SoccerNet 2023 challenges were the third annual video understanding challenges organized by the SoccerNet team. For this third edition, the challenges were composed of seven vision-based tasks split into three main themes. 
The first theme, broadcast video understanding, is composed of three high-level tasks related to describing events occurring in the video broadcasts: (1) action spotting, focusing on retrieving all timestamps related to global actions in soccer, (2) ball action spotting, focusing on retrieving all timestamps related to the soccer ball change of state, and (3) dense video captioning, focusing on describing the broadcast with natural language and anchored timestamps.
The second theme, field understanding, relates to the single task of (4) camera calibration, focusing on retrieving the intrinsic and extrinsic camera parameters from images.
The third and last theme, player understanding, is composed of three low-level tasks related to extracting information about the players: (5) re-identification, focusing on retrieving the same players across multiple views, (6) multiple object tracking, focusing on tracking players and the ball through unedited video streams, and (7) jersey number recognition, focusing on recognizing the jersey number of players from tracklets.
Compared to the previous editions of the SoccerNet challenges, tasks (2-3-7) are novel, including new annotations and data, task (4) was enhanced with more data and annotations, and task (6) now focuses on end-to-end approaches.
More information on the tasks, challenges, and leaderboards are available on \hyperlink{https://www.soccer-net.org}{https://www.soccer-net.org}. Baselines and development kits can be found on \hyperlink{https://github.com/SoccerNet}{https://github.com/SoccerNet}.$^*$Denotes equal contributions and $^\dag$the challenges organizers.
}




\maketitle

\section{Introduction}

The field of video understanding continues to captivate the focus of computer vision research. 
As part of the commitment to advance video analysis tools within the context of sports, the SoccerNet dataset has already introduced a set of eleven tasks related to video understanding. 
In 2023, seven of those tasks were part of yearly open challenges aimed at the broader research community. 
This paper describes the conclusive outcomes of the SoccerNet 2023 challenges, briefly showcasing the solutions proposed by the participants of each challenges. 
% providing space for participants to briefly showcase their solutions.

\subsection{SoccerNet dataset}

Originally introduced by Giancola~\etal~\cite{Giancola2018SoccerNet} in 2018, SoccerNet evolved into a substantial dataset tailored for reproducible research in soccer video understanding. The dataset initially had two objectives: offer a comprehensive benchmark for research in soccer video understanding and introduce the novel task of action spotting, focusing on the temporal localization of soccer actions among $500$ videos for three major actions: goals, cards, and substitutions.
Subsequent advancements emerged with the introduction of SoccerNet-v2 by Deliège~\etal~\cite{Deliege2021SoccerNetv2}, which expanded the annotations to a complete set of common soccer actions such as penalties, clearances, ball out of play, and more. SoccerNet-v2 also featured insights into camera transitions encompassing $13$ camera classes for the task of camera shot segmentation. If a camera shot featured an action replay, it was linked to its corresponding action timestamp during the live broadcast, which defined a replay grounding task.

In 2022, Cioppa~\etal~\cite{Cioppa2022Scaling} introduced SoccerNet-v3, including new spatial annotations covering players, the ball, field lines, and goal parts from various viewpoints of the same scene.  Alongside, three new tasks were proposed: pitch localization, camera calibration, and player re-identification. 
The same year, SoccerNet-Tracking~\cite{Cioppa2022SoccerNetTracking}, introduced the task of multiple object tracking across video clips, fostering long-term tracking and including metadata such as jersey numbers and team affiliations.

Finally, SoccerNet received two upgrades in 2023. The first one, SoccerNet-Captions~\cite{Mkhallati2023SoccerNetCaption}, introduced natural language descriptions of events in the broadcast games, defining a new dense video captioning task. The second one, SoccerNet-MVFouls~\cite{Held2023VARS}, introduced a multi-view dataset for foul recognition and characterization.

\subsection{SoccerNet challenges}

The 2023 edition of the SoccerNet challenges proposed a set of seven tasks related to soccer video understanding. They were grouped into three main themes, painting a comprehensive picture that spanned the majority of soccer video analyses. 
The first theme, \textbf{broadcast video understanding}, included three major high-level tasks: (1) action spotting, focusing on retrieving all timestamps related to global actions in soccer, (2) ball action spotting, focusing on retrieving all timestamps related to the soccer ball change of state, and (3) dense video captioning, focusing on describing the broadcast with natural language and anchored timestamps. 
The second theme, \textbf{field understanding}, is centered on the task of (4) camera calibration, focusing on retrieving the intrinsic and extrinsic camera parameters from images. 
The third and last theme, \textbf{player understanding}, introduced three tasks centered on players: (5) re-identification, focusing on retrieving the same players across multiple views, (6) multiple object tracking, focusing on tracking players and the ball through unedited video streams, and (7) jersey number recognition, focusing on recognizing the jersey number of players from tracklets.
Compared to previous editions of the SoccerNet challenges~\cite{Giancola2022SoccerNet}, tasks (2-3-7) are novel, introducing extra annotations and data. Task (4) was made richer with additional data and annotations, while task (6) changed its focus towards a more comprehensive approach.

%For a full grasp of these tasks, challenges, and the leaderboards, detailed information is available at \hyperlink{https://www.soccer-net.org.}{https://www.soccer-net.org.} Foundational starting points and kits for development, which play a crucial role in getting participants started, can be easily accessed at \hyperlink{https://github.com/SoccerNet}{https://github.com/SoccerNet}.

\subsection{Individual contributions}

Anthony Cioppa and Silvio Giancola are the lead organizers of the SoccerNet challenges. They are also the lead task organizers for the action spotting and ball action spotting challenges. Vladimir Somers is the lead task organizer of the tracking challenge with Xin Zhou, as well as the re-identification and jersey number recognition challenges. Floriane Magera is the lead task organizer for the camera calibration challenge. Hassan Mkhallati is the lead task organizer for the dense video captioning challenge. Adrien Deliège co-initiated the action spotting, camera calibration, tracking, and re-identification challenges. Jan Held and Carlos Hinojosa helped with the practical organization and communication of the challenges. Amir M. Mansourian formatted the jersey number recognition dataset. Pierre Miralles provided the ball action spotting dataset, including the videos and annotations. Olivier Barnich, Christophe De Vleeschouwer, Alexandre Alahi, Bernard Ghanem and Marc Van Droogenbroeck are the supervisors and provided funds for the annotations or the human resources to organize those challenges. The remainder of the authors are the participants who provided a summary of their method in the paper, listed in alphabetical order.

\subsection{Manuscript organization}

The manuscript is organized into seven sections, each summarizing the findings of a task.
For each section, we describe the task and the metric used to evaluate the participant. We highlight the leaderboard of our participants for this year's challenges, and the leader provides a summary of their method. We conclude each section with the significant findings for that task. The remainder of the summaries can be found in the appendix.

\section{Action Spotting}\label{sec:ActionSpotting}

\subsection{Task description}

Action spotting consists in localizing the exact moments when actions of interest occur, anchored by single timestamps (\eg a free-kick is defined by the precise moment when the player kicks the ball).
Similarly to the two past editions of this challenge~\cite{Giancola2022SoccerNet}, the dataset comprises $500$ games, with a total of $110{,}458$ actions spanning $17$ categories. In addition, a set of an extra $50$ games with segregated annotations is used as the challenge set.

\subsection{Metrics}

We use the Average-mAP~\cite{Giancola2018SoccerNet} metric to evaluate action spotting.
A predicted action spot is considered a true positive if it falls within a given tolerance $\delta$ of a ground-truth timestamp from the same class. 
The Average Precision (AP) based on Precision-Recall (PR) curves is computed then averaged over the classes (mAP), after which the Average-mAP is calculated as the AUC of the mAP at different tolerances $\delta$. 
The \textit{loose Average-mAP} uses tolerances $\delta$ ranging from $5$ to $60$ seconds~\cite{Giancola2018SoccerNet} and the \textit{tight Average-mAP} stricter tolerances $\delta$ ranging from $1$ to $5$ seconds.

%Moreover, we differentiate between actions that are visible in the broadcast video, versus the actions that are not directly shown. For instance, several throw-ins and indirect free-kicks are not shown in the broadcast but can still be inferred from the dynamic of a game, after a ball went out of play or after a foul occurred. Spotting unshown actions requires a more abstract level of understanding involving the learning of causality and game logic.

\subsection{Leaderboard}

This year, $10$ teams participated in the action spotting challenge for a total of $55$ submissions, with an improvement from $68.33$ to $71.31$ tight average mAP. The leaderboard may be found in Table~\ref{tab:LeaderboardSpotting}.

\subsection{Winner}
\label{sub:actionspottingwinner}

The winners for this task are Wenjie Zhang \etal. A summary of their method is given hereafter.

\mysection{A1 - MEDet: Multi-Encoder Fusion for Enhanced Action Spotting Task}\\
\textit{Wenjie Zhang, Ran Song, Ziyu Meng, Zhiheng Li, Tianjiao Li, and Wei Zhang}\\
\textit{\{zwjie, mziyu, zhihengli\}@mail.sdu.edu.cn, \{ransong, davidzhang\}@sdu.edu.cn, \\tianjiao\_li@mymail.sutd.edu.sg}
%\textit{zwjie@mail.sdu.edu.cn, ransong@sdu.edu.cn, \\mziyu@mail.sdu.edu.cn, \\zhihengli@mail.sdu.edu.cn, \\tianjiao\_li@mymail.sutd.edu.sg, davidzhang@sdu.edu.cn}

MEDet is designed for the task of detecting action instances in long untrimmed videos. We delve into the capabilities of convolutional neural networks (CNNs) and transformer neural networks in capturing local and global features and design three different encoders including Conv-based, Transformer-based and CNN-Transformer Hybrid architectures. To handle properties of various actions, MEDet divides the whole actions into three groups corresponding to different encoders. Furthermore, we proposed an improved feature pyramid network to obtain enhanced multi-scale features. Finally, our decoder utilizes a CNN-based classification head and a Trident regression head to obtain action labels and action boundaries. At inference, all result sets generated by three branches are merged and then filtered by the Non-Maximum Suppression (NMS) algorithm (in a way of model ensemble). Without using any additional datasets, the proposed MEDet demonstrates superior performance on the action spotting task.

\subsection{Results}

Even though this was the third edition of the action spotting challenge, we saw an improvement for $5$ out of the $10$ teams over last year's state-of-the-art results by Soares~\etal~\cite{Soares2022Temporally}.
Specifically, these teams proposed to use (i) different encoder branches for different types of actions that have similar dynamics, (ii) multi-scale pyramidal architectures, (iii) fusions of features including CLIP and Video MAE features, and (iv) self-supervised pre-training.
% Specifically, these teams proposed to use several strategies: the use of different encoder branches for different types of actions that have similar dynamics, the use of multi-scale pyramidal architectures, the fusion of features including CLIP and Video MAE features, and self-supervised pre-training.

% [TBD] Write paragraph from following information and discuss the leaderboard:
% \begin{itemize}
%     \item Different encoder branches for different types of actions
%     \item Use of multi-scale pyramidal architectures
%     \item Fusion of features with new ones such as CLIP and Video MAE
%     \item Self-supervised pre-training
% \end{itemize}

\begin{table}[t]
    \caption{Action spotting leaderboard. Main metric for the leaderboard and best performances in bold. Team names with a superscript have provided a summary that may be found in Appendix~\ref{app:spotting} or in Section~\ref{sub:actionspottingwinner} for the winning team.
    }
    \label{tab:LeaderboardSpotting}
    \centering
\resizebox{\columnwidth}{!}{%
    \begin{tabular}{l||c|c|c||c|c|c}
% Title
\multirow{ 2}{*}{Participants} & \multicolumn{3}{c||}{\textbf{tight Average-mAP}} & \multicolumn{3}{c}{loose Average-mAP} \\ %\cline{2-7}
& \textbf{main} & vis. & inv. & main & vis. & inv. \\ \midrule
% Teams
\bf{SDU\_VSISLAB$^{S1}$} & \bf{71.31} &  76.29  & 54.09 & 78.56  & 81.67 & 69.13 \\
mt\_player$^{S2}$ & 71.10 & \textbf{77.22}  & 58.50 & 78.79  & 82.02 & 77.62 \\
ASTRA$^{S3}$~\cite{Xarles2023ASTRA} &  70.10 & 75.00  & 57.98 & \textbf{79.21}  & \textbf{81.69} &  75.36 \\
team\_ws\_action & 69.17 & 75.18 & 59.12 & 76.95 & 80.39 & 75.92\\
CEA LVA$^{S5}$ &68.38 &74.79 &	47.68 &	73.98 & 78.57 &	61.75\\
Baseline \cite{Soares2022Temporally}* & 68.33 & 73.22  & \textbf{60.88} & 78.06  & 80.58 & \textbf{78.32} \\
DVP  &66.95 &	74.68 &	53.81 &	73.61 &	79.15 &	67.38 \\
JAMY2 (AF\_GRU) &	51.97 	&58.05 	&44.29 	&63.12 	&65.98 &	61.66 \\
tyru (GRU\_CALF) &	51.38 	&57.50 	&41.82 	&62.88 	&66.30 &	56.57 \\
JAMY (LocPoint) &	45.83 	&49.68 	&45.71 	&61.80 	&64.23 &	63.48 \\
test\_YYQ &	12.73 &	14.13 &	11.21 	&54.21 	&58.75 	&48.55 \\

    \end{tabular}%
}
\end{table}

\section{Ball Action Spotting}\label{sec:BallActionSpotting}

\subsection{Task description}

This novel task of ball action spotting consists in localizing the exact time when two types of actions related to the soccer ball occur: \textit{pass} and \textit{drive}. These actions are anchored with a single timestamp corresponding to the exact time the ball leaves a player's foot for a \textit{pass} event and when a player touches the ball to control it for a \textit{drive} event. The dataset is provided by Footovision and is composed of $7$ games with a total of $11{,}041$ annotated timestamps, as well as $2$ extra segregated games for the challenge set.

\subsection{Metrics}

Due to the fast nature of the event, we evaluate the performance of the methods based on the tight average mAP as well as the mAP at different $\delta$ thresholds ranging from $1$ to $5$ seconds. We choose to rank the methods with respect to the mAP@$1$, meaning that participants need to localize all ball events very precisely.

\subsection{Leaderboard}

This year, $5$ teams participated in the action spotting challenge for a total of $102$ submissions, with an improvement from $62.72$ to $86.47$ average mAP@$1$. The leaderboard may be found in Table~\ref{tab:LeaderboardBallSpotting}.

\subsection{Winner}
\label{sub:ballactionspottingwinner}

The winner of this task is Ruslan Baikulov. A summary of his method is given hereafter.

\mysection{B1 - Ruslan Baikulov}\\
\textit{Ruslan Baikulov} \textit{ruslan1123@gmail.com}

The model architecture and multi-stage training significantly contribute to the overall metric outcome of the solution.
The architecture utilizes a slow fusion approach, incorporating 2D and 3D convolutions. The model consumes sequences of grayscale frames stacked in sets of three. The shared 2D encoder independently processes these input threes, producing visual features. The following 3D encoder processes visual features, producing temporal features. Then, a linear classifier predicts the presence of actions.
Multi-stage training was carried out in four steps. The first stage is basic training with the 2D encoder initialized with pre-trained ImageNet weights. The second stage is the same training but on the Action Spotting Challenge dataset. The third stage uses predictions from the first stage for hard negative sampling and encoders weights from the second stage for initialization. The fourth stage is fine-tuning weights from the third stage on long sequences (33 frames instead of 15 before).

\subsection{Results}

This new task brings three novel difficulties to the realm of action spotting. First, it focuses on fast and subtle events that provide few visual cues compared to the overall video. Second, the events are much denser compared to the action spotting challenge. It is, therefore, hard to differentiate between two actual close events or a wrong double detection.  Finally, the amount of provided data is small compared to the action spotting challenge, encouraging participants to try semi-supervised, self-supervised, or transfer learning paradigms using the $500$ broadcast games.

The participants focused on several aspects for this first edition of the challenge. 
First, some participants showed they could improve their performance by performing pre-training on the action spotting videos and later fine-tuning the network on the ball action spotting task. 
Next, stacked sequences of grayscale images in the RGB channels and label expansion with focal loss helped improve the performance. 
Finally, model ensembling helped improve the performance further by generating several network variants.

%[TBD] Write paragraph from following information and discuss the leaderboard:
%\begin{itemize}
%    \item False RGB images using concatenated grayscale images.
%    \item Pre-training on SoccerNet Action Spotting and fine-tuning.
%    \item Model ensembling by generating variants.
%    \item Label expansion and focal loss.
%    \item Fusion of Motion, RGB, Grayscale, and ball crops.
%    \item Bottom-up approach based on object detection.
%\end{itemize}

\begin{table}[t]
    \caption{Ball Action spotting leaderboard. The main metric for the leaderboard and best performances are in bold. Team names with a superscript have provided a summary that may be found in Appendix~\ref{app:ballspotting} or in Section~\ref{sub:ballactionspottingwinner} for the winning team.
    }
    \label{tab:LeaderboardBallSpotting}
    \centering
\resizebox{\columnwidth}{!}{%
    \begin{tabular}{l||c|c|c|c|c||c}
% Title
\multirow{ 2}{*}{Participants} & \multicolumn{5}{c||}{\textbf{mAP}} & \multicolumn{1}{c}{Average-mAP} \\ %\cline{2-7}
&  \textbf{@1} & @2 & @3 & @4 & @5 & tight \\ \midrule
% Teams
\textbf{Ruslan Baikulov}$^{B1}$ 	&\textbf{86.47} 	&\textbf{87.98} 	&\textbf{88.28} 	&\textbf{88.18} 	&\textbf{87.95} 	&\textbf{87.91}\\
FDL@ZLab$^{B2}$ 	&83.39 	&85.19 	&85.81 	&86.00 	&86.19 	&85.45\\
BASIK$^{B3}$	&82.06 	&83.39 	&83.86 	&84.04 	&83.91 	&83.57\\
FC Pixel Nets$^{B4}$ 	&81.89 	&83.22 	&83.97 	&83.85 	&84.02 &	83.50\\
play 	&79.74 	&82.58 	&84.06 	&84.49 	&84.34 	&83.29\\
Baseline~\cite{Hong2022Spotting}* 	&62.72 	&69.24 	&72.57 	&74.29 	&74.80 	&71.21\\

    \end{tabular}%
}
\end{table}

\section{Dense Video Captioning}\label{sec:DenseVideoCaptioningwinner}

\subsection{Task description}

Dense video captioning is a new task introduced by Mkhallati~\etal~\cite{Mkhallati2023SoccerNetCaption}. Given a long untrimmed video, the task consists in spotting all instants where a comment should be anchored and generating sentences describing the events occurring around that time using engaging natural language.
The SoccerNet-Caption dataset comprises $471$ untrimmed broadcast games at 720p resolution and 25fps.
This first edition of the challenge focuses on the anonymized comments provided with the dataset, for a total of $36{,}894$ timestamped comments. In addition, a set of $42$ extra games with segregated annotations is used as the challenge set.

\subsection{Metrics}

Established captioning evaluation metrics such as METEOR~\cite{Lavie2007Meteor}, BLEU~\cite{Papineni2001BLEU}, ROUGE~\cite{Lin2004ROUGEAP}, and CIDEr~\cite{Vedantam2015CIDEr} are adapted to estimate the language similarity between all generated captions with any ground-truth caption for which its timestamps fall within a $\delta$ tolerance. Then, the performances are averaged over the video and the dataset. We choose the METEOR@$30$, corresponding to a time tolerance of $30$ seconds around the ground-truth captions, as the primary ranking metric for this challenge. %As a secondary metric, we also use an adapted version of the SODA\_c~\cite{Fujita2020SODA} metric.

\subsection{Leaderboard}

This year, $4$ teams participated in the camera calibration challenge for a total of $34$ submissions, with an improvement from $21.25\%$ to $26.66\%$ METEOR@$30$. The leaderboard may be found in Table~\ref{tab:LeaderboardCaption}.

\subsection{Winner}
\label{sub:densevideocaptioning}

The winners for this task are Zheng Ruan~\etal. A summary of their method is given hereafter.

\mysection{D1 - OPPO}\\
\textit{Zheng Ruan, Ruixuan Liu, Shimin Chen, Mengying Zhou, Xinquan Yang, Wei Li, Chen Chen, and Wei Shen}\\
\textit{\{liuruixuan, chenshimin1, yangxinquan, liwei19, chenchen, shenwei12\}@oppo.com
\{2130730, 2130904\}@tongji.edu.cn
}

For video captioning, we modified Blip~\cite{Li2022BLIP-arxiv} as our framework. We pick a second every 10 seconds and select the first
frame of each of the 16 seconds before and after that second as input. We apply ViT~\cite{Dosovitskiy2020AnImage-arxiv} as the vision encoder to extract
features. To better use spatial-temporal information, we add a perceiver resampler~\cite{Alayrac2022Flamingo-arxiv} after the vision encoder, which can map
spatial-temporal visual features to a fixed length learned latent vectors. Then we use BERT~\cite{Devlin2019Bert} as the vision-grounded text
decoder for captions generation. For localization, we use 0.875 as the threshold to filter the low-quality caption. When the
confidence of the output caption is higher than the threshold, the caption will be selected. Compare with other methods~\cite{Alayrac2022Flamingo-arxiv,Li2023BLIP2-arxiv}
our framework performs the best on both CIDEr and Meteor. After adding filtering, the performance is improved by +5.124\%
in Meteor, which shows the effectiveness of filtering.
References

\subsection{Results}

As this was the first edition of the challenge on a novel task officially presented on arXiv in April 2023, only a few teams were able to provide results on the challenge set in time for the challenge. Nevertheless, two teams were able to improve the performance compared to our second baseline presented in the work of Mkhallati~\etal~\cite{Mkhallati2023SoccerNetCaption}. 
Specifically, these methods leveraged pre-trained video-language transformers and ensembling methods to select the most suitable generated captions.
As one can see from Table~\ref{tab:LeaderboardCaption}, the winning team significantly improved on all metrics compared to the rest of the methods, especially for the CIDEr@$30$ metric. 

% [TBD] Write paragraph from following information and discuss the leaderboard:
% \begin{itemize}
%     \item Pre-trained video-language transformers
%     \item Ensemble method to select the generated caption 
%     \item De-duplication and filtering
% \end{itemize}

\begin{table}[t]
    \caption{Dense video captioning leaderboard. The main metric for the leaderboard and best performances are in bold. 
    % The winning team names provided a summary in Section~\ref{sub:densevideocaptioning}.
    Team names with a superscript have provided a summary that may be found in Appendix~\ref{app:spotting} or in Section~\ref{sub:densevideocaptioning} for the winning team.
    }
    \label{tab:LeaderboardCaption}
    \centering
\resizebox{\columnwidth}{!}{%
    \begin{tabular}{l||c|c|c|c|c|c|c}
% Title
\multirow{ 2}{*}{Participants} & \multicolumn{7}{c}{\textbf{Metric@30}}  \\ %\cline{2-7}
& \textbf{METEOR} & BLEU\_1 & BLEU\_2 & BLEU\_3 & BLEU\_4 & ROUGE\_L & CIDEr \\ \midrule
% Teams
\textbf{OPPO}$^{D1}$ 	&\textbf{26.66} 	&\textbf{35.55} 	&\textbf{31.03}	&\textbf{28.13} 	&\textbf{25.65} 	&\textbf{33.23} 	&\textbf{69.73} 	\\
HZC 	&21.30 	&29.73 	&24.52 	&21.44 	&19.13 	&24.56 	&24.76 \\
Baseline$_2$* 	&21.25 	&30.01 	&24.80 	&21.74 	&19.44 	&24.65 	&25.68\\
justplay$^{D3}$ 	&21.20 	&29.83 &	24.68 &	21.66 &	19.38 &	24.34 &	25.89\\
aisoccer 	&21.02 &	29.53& 	24.42& 	21.42& 	19.15& 	24.31& 	23.72\\
Baseline$_1$* 	&15.24 &	11.91& 	9.97 &	8.83 	&7.97 	&10.69 	&16.33 \\

    \end{tabular}%
}
\end{table}

\section{Camera Calibration}\label{sec:CameraCalibration}

\subsection{Task description}

Camera calibration consists in estimating the intrinsic and extrinsic camera parameters based on an image. Similarly to last year's challenge, the pinhole camera model with optional distortion (tangential, radial, and thin prism) is imposed.
This year's dataset was complemented with extra images and annotations compared to the 2022 edition. Specifically, the dataset contains $25{,}506$ images and $226{,}305$ annotated polylines. In addition, a set of $2{,}690$ images with segregated annotation is used as the challenge set. 

\subsection{Metrics}

Since no camera parameter ground truths are available for our soccer images, the methods are evaluated using the re-projection error with the manually annotated field lines in the 2D image plane. 
Compared to last year's challenge, we simplified the ranking metric (named \textit{combined metric}) by computing it as the multiplication of the \textit{accuracy@$5$} and the \textit{completeness score}. All details on how to compute those two metrics are given in Section $5$ of Giancola~\etal~\cite{Giancola2022SoccerNet}.

%First, the provided 3D soccer pitch model is projected on the 2D image plane using the estimated camera calibration parameters.
%Since some images are challenging to calibrate, we let the participants the opportunity to skip some images. This defines a completeness score which is the ratio of images for which the camera parameters are provided over the total number of images in the evaluation set.

\subsection{Leaderboard}

This year, $7$ teams participated in the camera calibration challenge for a total of $66$ submissions, with an improvement from $8\%$ to $55\%$ for the combined metric. The leaderboard may be found in Table~\ref{tab:LeaderboardCalibration}.

\subsection{Winner}
\label{sub:cameracalibrationwinner}

The winners for this task are Nikolay Falaleev~\etal. A summary of their method is given hereafter.

\mysection{C1 - Sportlight}\\
\textit{Nikolay Falaleev and Ruilong Chen}\\
\textit{nikolay.falaleev@sportlight.ai, \\ruilong.chen@sportlight.ai}

Our solution was based on a combination of keypoint and line detections. In total, there were 57 keypoints, most of which were defined by intersecting fitting results of lines and ellipses from annotations. To increase the overall number of available points, we utilized the correspondence between tangent points of the circles to add 8 tangent points. Additionally, 13 points were added through homography projection.
We used an HRNetV2-w48-based neural network for detecting the keypoints as heatmaps, which have peaks at the keypoints locations. Similarly, 23 lines were detected, and each line was represented by a heatmap with two peaks indicating the location of the line's extremities.
The camera parameters were determined using keypoints and line intersections. We applied the standard OpenCV camera calibration algorithm to subsets of points selected with various heuristics. The final camera parameters were chosen through a heuristic voting mechanism based on the reprojection error.

\subsection{Results}

This year, the participants improved their method by following three main research directions. 
First, some participants detected new pitch elements such as circles and lines intersections which improved their pitch element localization and therefore the derived camera calibration parameters. 
Second, some participants used keypoints spread uniformly on the ground to improve the estimated camera parameters. 
Finally, we saw the first solution based on differential rendering, optimizing directly the camera parameters given the synthetic projection of pitch zones and image segmentations.

% \subsection{Results}
% [TBD] Write paragraph from following information and discuss the leaderboard:
% \begin{itemize}
%     \item Detection of new pitch elements circle and lines intersections
%     \item keypoints spread uniformly on the ground 
%     \item First solution based on differential rendering optimizing directly camera parameters given the synthetic projection of pitch zones and the image segmentation.
% \end{itemize}

\begin{table}[t]
\caption{Camera calibration leaderboard. The main metric for the leaderboard and best performances are in bold. Team names with a superscript provided a summary that can be found in Appendix \ref{app:cameracalibration}, or in Section \ref{sub:cameracalibrationwinner} for
the winner. %The baseline description may be found in \url{https://github.com/SoccerNet/sn-calibration}.
}
\label{tab:LeaderboardCalibration}

\resizebox{\columnwidth}{!}{%
\begin{tabular}{l||c||c|c}
Participants & \textbf{Combined} & ACC@5 & Completeness\\ 
 \midrule
Sportlight$^{C1}$ 	&\textbf{0.55} &\textbf{73.22} &	75.59\\
Spiideo$^{C2}$ &0.53 	&52.95 	&\textbf{99.96}\\
SAIVA\_Calibration$^{C3}$ &0.53 &60.33 &87.22\\
BPP &	0.50 	&69.12 	&72.54\\
ikapetan &	0.43 &	53.78& 	79.71\\
NASK$^{C6}$ &	0.41 &	53.01 	&77.81\\
MA \& JT 	&0.41 	&58.61 	&69.34\\
Baseline* 	&0.08 	&13.54 	&61.54
% \hline 
\end{tabular}%
}
\end{table}

\section{Player Re-Identification}\label{sec:PlayerRe-Identification}

\subsection{Task description}
Person re-identification \cite{Ye2022DeepLF}, or simply ReID, is a person retrieval task that aims at matching an image of a person-of-interest, called the \textit{query}, with other person images within a large database, called the \textit{gallery}, captured from various camera viewpoints. 
Re-identification is a challenging task, because person images generally suffer from background clutter, inaccurate bounding boxes \cite{Zheng2015PartialPR}, luminosity changes \cite{Wei2017PersonTG}, and occlusions \cite{Somers2023Body} from street objects or other people.
The goal of the SoccerNet ReID task is to re-identify players and referees across multiple camera viewpoints for a given action at a specific time instant during a soccer game.
Our SoccerNet re-identification dataset is composed of $340{,}993$ players' thumbnails extracted from image frames of broadcast videos from $400$ soccer games within $6$ major leagues.
Compared to traditional street surveillance-type re-identification datasets, the SoccerNet-v3 ReID dataset is particularly challenging because soccer players from the same team have very similar appearances, which makes it difficult to tell them apart. 
On the other hand, each identity has a few amount of samples, which makes the model even more difficult to train. 
Finally, there is a big diversity within samples of the dataset in terms of image resolution.

\subsection{Metrics}
We use two standard retrieval evaluation metrics to compare different ReID models: the cumulative matching characteristics (CMC) \cite{Wang2007ShapeAA} at Rank-1 and the mean average precision \cite{Zheng2015ScalablePR} (mAP).
The participants in the 2023 SoccerNet re-identification challenge are ranked according to their mAP score on a segregated challenge set.

\subsection{Leaderboard}

This year, $5$ teams participated in the re-identification challenge, for a total of $28$ submissions, with an improvement from $91.68\%$ to $93.26\%$ mAP. The leaderboard may be found in Table~\ref{tab:LeaderboardReID}.

\subsection{Winner}
\label{sub:playerreidentificationwinner}

The winners for this task are Konrad Habel~\etal. A summary of their method is given hereafter.

\mysection{R1 - CLIP-ReIdent: Contrastive Training for Player Re-Identification}\\
\textit{Konrad Habel, Fabian Deuser, and Norbert Oswald}\\
\textit{konrad.habel@unibw.de, fabian.deuser@unibw.de, norbert.oswald@unibw.de}

Our approach is mainly based on our paper CLIP-ReIdent: Contrastive Training for Player Re-Identification~\cite{Habel2022CLIPReIdent}. This approach is also the $1^{st}$ place winning solution for the Player Re-Identification challenge 2022~\cite{VanZandycke2022DeepSportradarv1} at the ACM Multimedia MMSports Workshop. For the SoccerNet challenge we use an ensemble of three CLIP-based models from OpenCLIP~\cite{Ilharco2021OpenCLIP} and OpenAI~\cite{Radford2021Learning}. Our approach utilizes a custom sampling strategy during training to sample players of the same action together. Furthermore, we use a self-designed re-ranking per action as post-processing. For the ensemble the vision encoders of the CLIP models are fine-tuned with contrastive training and InfoNCE loss as training objective on the data of the Train and Validation set. Our solution achieves on the Test set a mAP of $ 93.51\%$ and on the Challenge set $93.26\%$.

\subsection{Results}
Last year, participants came up with various innovative ideas and achieved outstanding performances despite the difficulty of the task. 
Last year's winning solution was used as the baseline, but this year's winning team managed to improve upon it and set a new state-of-the-art performance.
This year's solutions were mostly based on ViT \cite{Dosovitskiy2020AnII}, and most teams adopted the foundation model CLIP \cite{Radford2021Learning}. Moreover, participants used model ensembling, per-action player re-ranking techniques, and custom action-based training samplers to improve ranking results.

\begin{table}[t]
\caption{Re-identification leaderboard. The main metric for the leaderboard and best performances are in bold. The winning team provided a summary that can be found in Section~\ref{sub:playerreidentificationwinner}.}
\label{tab:LeaderboardReID}
\centering
 \resizebox{\columnwidth}{!}{%
\begin{tabular}{l||c|c@{\hspace*{0.3cm}}|@{\hspace*{0.08cm}}|@{\hspace*{0.3cm}}l||c|c}
% Title
Participants & \textbf{mAP} & R-1 & Participants & \textbf{mAP} & R-1 \\ \midrule
% Teams
\textbf{UniBw Munich - VIS$^{R1}$} 	&\textbf{93.26} 	&\textbf{91.26} &MTVACV 	&90.11 	&87.04\\
Baseline (Inspur)* 	&91.68 	&89.41 &ErxianBridge 	&85.76 	&82.33\\
sjtu-lenovo 	&91.51 	&89.17 &cm\_test 	&42.60 	&28.73\\
\end{tabular}%
 }

\end{table}

\section{Multiple Player Tracking}\label{sec:MultiplePlayerTracking}

\subsection{Task description}
The task of multiple player tracking aims to track individual subjects (\eg players or ball) across frames. It is useful for downstream applications such as player highlights and player statistics.
Different from last year's challenge~\cite{Giancola2022SoccerNet}, we did not provide any ground truth bounding boxes for the objects to track, making this task more challenging. Instead, the participants have to solve both detection and association. 
% 
% In last year's challenge~\cite{Giancola2022SoccerNet}, we split the tracking task into two steps: (1) detecting the objects to track and (2) associating the bounding boxes over time to create the tracklets and provided ground-truth bounding boxes for step (1). In this year's challenge, no ground-truth detection were provided, so the resulting solution has to solve both detection and association. 
%
Compared to most tracking datasets, re-identification present extra challenges since player appearances are very similar, fast moving, and may leave the frames and come back.

\subsection{Metrics}

We keep the HOTA metric to rank the participants as proposed in Luiten~\etal~\cite{Luiten2021Hota}. The metric combines a detection accuracy (DetA) and an association accuracy (AssA).

\subsection{Leaderboard}

This year, $7$ teams participated in the multiple player tracking challenge for a total of $83$ submissions, with an improvement from $42.38\%$ to $75.61\%$ HOTA. 
Table~\ref{tab:LeaderboardTracking} presents the leaderboard. % The leaderboard may be found in Table~\ref{tab:LeaderboardTracking}.

\subsection{Winner}
\label{sub:multipleplayertrackingwinner}

The winners for this task are Adrien Maglo~\etal. A summary of their method is given hereafter.

\mysection{T1 - Kalisteo}\\
\textit{Adrien Maglo, Astrid Orcesi, Quoc-Cuong Pham}\\
\textit{adrien.maglo@cea.fr, astrid.orcesi@cea.fr, Quoc-Cuong.pham@cea.fr}

After having detected the players with a YOLO-X model, TrackMerger v2
generates player tracklets by sequentially processing the video
frames. The current frame detections are matched to existing tracklets
bounding boxes with two successive Hungarian assignment algorithms. The
Intersection-Over-Union between bounding boxes and the distance
between their center are used as criteria. A Kalman filter with camera
motion compensation predicts the positions of existing tracklets in the
current frame.
The generated tracklets are subsequently split if they cross each other
to remove most association errors.
These non-ambiguous tracklets are used to fine-tune a Multiple
Granularity Network re-identification model with a triplet loss
formulation. Positive samples are extracted from the same tracklets as
the anchor while negative samples come from concomitant tracklets. To
generate full tracks, tracklets are iteratively merged according to the
distance between their re-identification vectors, preventing
player duplication and teleportation.

\begin{table}[t]
    \caption{Tracking leaderboard. The main metric for the leaderboard and best performances are in bold. Team names with a superscript have provided a summary that may be found in Appendix~\ref{app:tracking}, or in Section~\ref{sub:multipleplayertrackingwinner} for the winners.}
    \label{tab:LeaderboardTracking}
    
    \centering
\resizebox{\columnwidth}{!}{%
    \begin{tabular}{l||c|c|c}
% Title
Participants & \textbf{HOTA} & DetA & AssA \\ \midrule
% Teams
\textbf{Kalisteo$^{T1}$} 	&\textbf{75.61} 	&\textbf{75.38} 	&\textbf{75.94}\\
MTIOT$^{T2}$ 	&69.54 	&75.18 	&64.45\\
MOT4MOT$^{T3}$\cite{Shitrit2023SoccerNet-arxiv} 	&66.27 	&70.32& 	62.62\\
ICOST$^{T4}$ 	&65.67 	&73.07 &	59.17\\
SAIVA\_Tracking$^{T5}$ &	63.20 	&70.45 &	56.87\\
ZTrackers$^{T6}$ 	&58.69&68.69 	&50.25\\
scnu 	&58.07 	& 64.77 	&52.23\\
Baseline* 	&42.38 &	34.41 &	52.21\\
    \end{tabular}%
}
\end{table}

\subsection{Results}
For the detection step, most participants fine-tuned the YoloX detector. 
The maturity of the Yolo framework and its lightweight made it the go-to choice, allowing to achieve a DetA close to $75$. 
Since no ground-truth detections were provided, lower detection scores affected the association step.
As the association step was more difficult, tackling these challenges contributed to increase the AssA score.
% There are a few difficulties in the association step which contributes additively to increasing the AssA
% As mentioned in the task description, there are a few difficulties in the association step. 
% Hence, handling more of them contributes additively to increasing the AssA score. 
More specifically, multiple participants went beyond simple position prediction methods to compensate for fast player motions, camera motions, and deblurring. 
Both two-stage (such as ByteTrack, SORT based) and end-to-end tracking methods achieved strong performance.
From the end-result point of view, the proposed methods perform much better than off-the-shelf open-source baselines. However, they are still some room for improvement in both detection and association.
% , leaving a good amount of space for improvement in both detection and association.

\section{Jersey Number Recognition}\label{sec:JerseyNumberRecognition}

\subsection{Task description}

The task consists in identifying the jersey number of a player from a short video tracklet showing the soccer players.
% Given short video tracklets of soccer players of a few hundred frames long, the participants have to identify the jersey number of each player. 
This jersey number recognition task is challenging because of the low quality of the thumbnails (\ie low resolution and high motion blur) and because the jersey numbers might be visible on a minimal subset of the whole tracklet.
% Players with jersey numbers that are not visible at all are annotated with the "-1" value.
The SoccerNet Jersey Number dataset comprises $2{,}853$ tracklets of players extracted from the SoccerNet tracking videos. 
The challenge set comprises $1{,}211$ separate players' tracklets with segregated annotations.
% The problem is framed as a multi-class classifier, with the classes being the jersey numbers ranging from $0$ to $99$.
% An extra class "$-1$" is defined when no jersey number is visible in the tracklet.
The target classes are therefore all the jersey numbers from 1 to 99, and one extra "$-1$" class when no jersey number is visible in the tracklet.

\subsection{Metrics}
Since the jersey number recognition challenge was formulated as a classification task, with one class for each possible jersey number in the $[0, 99]$ range, we use the overall classification accuracy as a target metric to rank participants' solutions. Therefore, we compute the jersey number prediction accuracy as the number of correctly predicted jersey numbers (including $-1$ for non-visible numbers) over the total number of tracklets in the challenge set.

\subsection{Leaderboard}

This year, $15$ teams participated in the jersey number recognition challenge, for a total of $157$ submissions, with the best method reaching $92.85\%$ accuracy. The leaderboard may be found in Table~\ref{tab:LeaderboardJersey}.

\begin{table}[t]
\caption{Jersey Number Recognition leaderboard. The main metric for the leaderboard and best performance are in bold. Team names with a superscript have provided a summary that may be found in Appendix~\ref{app:jerseynumber}, or in Section~\ref{sub:jerseynumberrecognitionwinner} for the winning team.}
\label{tab:LeaderboardJersey}
\centering
\resizebox{\columnwidth}{!}{%
\begin{tabular}{l|c||l|c}
% Title
Participants & \textbf{acc} & Participants & \textbf{acc}  \\ \midrule
% Teams
\textbf{ZZPM$^{J1}$} 	&\textbf{92.85} &Kalisteo 	&58.35\\ 
UniBw Munich - VIS$^{J2}$	&90.95 &FindNum 	&54.91\\ 
zzzzz$^{J3}$ 	&88.08 &jn 	&47.55\\ 
Mike Azatov 	&82.05 &Surya 	&28.40\\ 
MT-IOT$^{J5}$ 	&81.70 &tony506672558 	&20.06\\ 
justplay$^{J6}$ 	&77.77 &lfriend 	&5.68\\ 
AIBrain Global Team$^{J7}$ 	&75.18& zhq 	&4.07\\ 
SARG UWaterloo 	&73.77& Baseline (Random)* 	&3.93\\ 
\end{tabular}%
}

\end{table}

\subsection{Winner}
\label{sub:jerseynumberrecognitionwinner}

The winners for this task are Rui Peng~\etal. A summary of their method is given hereafter.

\mysection{J1 - ZZPM}\\
\textit{Rui Peng, Kexin Zhang, Junpei Zhang, Yanbiao Ma, and Licheng Jiao}\\
\textit{\{22171214876,22171214672,22171214671, ybmamail\}@stu.xidian.edu.cn, lchjiao@mail.xidian.edu.cn}

The task requires the identification of players' numbers, which become difficult due to high motion blur and low quality. Our solution was to first perform text detection on the training set, filtering out a portion of the data that clearly did not have numbers, using a modified pre-trained DBNet++ model for the training set, and removing images from the training set that did not contain a detection box with a confidence level of 90 or higher. Data augmentation was then performed on the dataset, including image rotation and flipping, image scaling and cropping, color scrambling, noise addition, and multi-frame image overlay methods. We found that the data augmentation method of multi-frame fusion can provide more information and increase the robustness to features such as number shape, color and texture, thus improving the accuracy of recognition. The enhanced data were then trained for text recognition using multiple models, including SVTR-tiny, SVTR-small, SATRN, NRTR, and ASTER. The final result consists of a fusion of votes from multiple models.

\subsection{Results}
Most teams adopted a standard three-stage approach. A first text detection step is performed with various state-of-the-art methods (DBNet++, MMOCR, Deepsolo, YOLO) to predict the jersey number location in the image. A second text recognition step is performed to recognize the corresponding number using fine-tuned state-of-the-art OCR methods (PP-OCRv3, PaddleOCR). Finally, majority voting is employed to aggregate image-level results within a tracklet and output the final video-level prediction.

% Some other original solutions were proposed to deviate from this standard pipeline: a video transformer network for end-to-end tracklet-level prediction, a super resolution network as pre-processing to enhance image quality, or a custom jersey number detection network based on the output of a pose estimation model and some heuristic leveraging human body priors about back torso location.

% [TBD] Write paragraph from following information and discuss the leaderboard:
% \begin{itemize}
%     \item Text detection with various methods (DBNet++, MMOCR, Deepsolo, YOLO, ...)
%     \item Text recognition with SotA OCR methods (PP-OCRv3, PaddleOCR, ...)
%     \item Image level prediction, then tracklet-level majority voting
%     \item Multi-task classification to train OCR
%     \item Video transformer network for tracklet-level prediction
%     \item Super resolution as pre-processing
%     \item Jersey number localisation based on pose estimation
% \end{itemize}

\section{Conclusion}\label{sec:conclusion}

This paper summarizes the outcome of the SoccerNet 2023 challenges. 
In total, we present the results on seven tasks: action spotting, ball action spotting, dense video captioning, camera calibration, player re-identification, player tracking, and jersey number recognition. 
These challenges provide a comprehensive overview of current state-of-the-art methods within each computer vision task. 
For each challenge, participants were able to significantly improve the performance over our proposed baselines or previously published state of the art.
Some tasks such as action spotting or player re-identification are reaching promising results for industrial use, while novel tasks such as dense video captioning may still require further investigation.
In the future, we will keep on extending the set of tasks, challenges, and benchmarks related to video understanding in sports.

\mysection{Acknowledgement.}
A. Cioppa is funded by the F.R.S.-FNRS.
This work was partly supported by the King Abdullah University of Science and Technology (KAUST) Office of Sponsored Research through the Visual Computing Center (VCC) funding and the SDAIA-KAUST Center of Excellence in Data Science and Artificial Intelligence (SDAIA-KAUST AI). 

\section*{Declarations} 
\mysection{Availability of data and code.} The data and code are available at these addresses \url{https://github.com/SoccerNet} \url{https://www.soccer-net.com}.

%\mysection{Ethics approval.} .

\mysection{Conflict of interest.} The authors declare no conflict of interest.

\mysection{Open Access.}%This article is licensed under a Creative Commons Attribution 4.0 International License, which permits use, sharing, adaptation, distribution and reproduction in any medium or format, as long  as you give appropriate credit to the original author(s) and the source,  provide a link to the Creative Commons licence, and indicate if changes were made. The images or other third-party material in this article are included in the article's Creative Commons licence, unless indicated otherwise in a credit line to the material. If material is not included in the article's Creative Commons licence and your intended use is not permitted by statutory regulation or exceeds the permitted use, you will need to obtain permission directly from the copyright holder. To view a copy of this licence, visit \url{http://creativecommons.org/licenses/by/4.0/}.


\bibliography{bib/abbreviation-short,
bib/abbreviation-short-modif,
bib/activity,
bib/bib,
bib/dataset-modif,
bib/labo-modif,
bib/new-refs-ok,
bib/soccer-modif,
bib/soccernet-challenge-modif,
bib/sports-modif,
bib/put-new-refs-here}

\appendix
\section{Appendix}
\label{sec:appendix}
In this appendix, the participants provide a short summary of their methods. Only teams who provided a technical report at the end of the challenge that has been peer-reviewed by the organizers were able to submit a summary. This ensures that the presented methods followed the challenge rules.

\subsection*{Action Spotting} \label{app:spotting}

\mysection{S2 - mt\_player}\\
\textit{Yingsen Zeng, Yujie Zhong, Zhijian Huang, Feng Yan, Lin Ma}\\
\textit{\{zengyingsen, zhongyujie, yanfeng05\} @meituan.com, huangzhj56@mail2.sysu.edu.cn, linma@alumni.cuhk.net}

Our proposed method aims to improve the accuracy of action detection in untrimmed videos through multi-scale and multi-feature fusion. The model is based on an encoderdecoder structure, including a fully convolutional encoder, a multi-scale feature pyramid network, and a lightweight decoder for action classification and location. To tackle single timestamp labeling for action locations, we employ a soft-NMS approach based on Euclidean distance after IoU-based NMS. Additionally, we discover that different video features exhibit biases towards different action categories, and therefore, we explore three methods of feature fusion: early fusion, mid-level fusion, and late fusion. Four pre-computed video features we use are Baidu, ResNet (both from Soccernet Challenge), CLIP~\cite{Radford2021Learning}, and VideoMAE~\cite{Wang2023VideoMAE}. Ultimately, we achieve the best performance by combining mid-level fusion and class-wise late fusion. Our method is extensively tested and demonstrates its effectiveness, achieving 71.1\% tight-mAP and 78.8\% loose-mAP in challenge set.

\mysection{S3 - ASTRA~\cite{Xarles2023ASTRA}}\\
\textit{Artur Xarles, Sergio Escalera, Thomas B. Moeslund,  Albert Clapés}\\
\textit{arturxe@gmail.com, sescalera@ub.edu, tbm@create.aau.dk, aclapes@ub.edu}

Action Spotting TRAnsformer (ASTRA) leverages pre-computed visual embeddings from five video classification backbones provided by Baidu Research~\cite{Zhou2021Feature-arxiv}. ASTRA combines these embeddings in a transformer encoder-decoder module with learnable queries on the decoder side. This design enables the model to handle different input and output temporal dimensions, resulting in improved performance with higher temporal resolution for the outputs. To further enhance ASTRA's capability to capture fine-grained details, we introduce the concept of temporally local attention within the transformer encoder. Following Soares et al.~\cite{Soares2022Temporally}, two prediction heads are employed to predict the action classification and the displacement offsets with respect to the predefined anchors. To enhance generalization and address the long-tail distribution of the data, ASTRA incorporates a balanced mix-up technique. This technique generates data mixtures during training using an action-balanced data distribution stored in a queue, which is updated during each batch iteration. More information is available in our paper published at MMSports'23~\cite{Xarles2023ASTRA}.

\mysection{S5 - COMEDIAN: Long-Context Transformer Pretraining Through Spatio-Temporal Knowledge Distillation for Action Spotting}\\
\textit{Julien Denize, Mykola Liashuha, Jaonary Rabarisoa, Astrid Orcesi, Romain Hérault}\\
\textit{\{julien.denize, mykola.liashuha, jaonary.rabarisoa, astrid.orcesi\}@cea.fr, romain.herault@insa-rouen.fr}

COMEDIAN is an approach to pretrain spatiotemporal transformer architectures to enrich local spatiotemporal information with a larger context before specializing in the action spotting task. These transformer architectures contain two encoders. The first is spatial as it takes input frames from a short video to embed their information in one output token. The second encoder is a temporal encoder that takes as input the spatial output tokens of sub-videos from a large video. To pretrain the architecture, knowledge distillation is performed between the output tokens of the temporal encoder and their provided temporally aligned Baidu features~\cite{Zhou2021Feature-arxiv}. The architecture is then finetuned to the action spotting task by performing the classification of each action independently on each timestamp associated with the temporal output tokens. For inference, a sliding window is performed on videos to provide predictions of each action at each timestamp followed by a soft NMS per class.

\subsection*{Ball Action Spotting} \label{app:ballspotting}

\mysection{B2 - Boosted Model Ensembling (BME)}\\
\textit{Luping Wang, Hao Guo, and Bin Liu}\\
\textit{\{wangluping, guoh, liubin\}@zhejianglab.com}

Our method, Boosted Model Ensembling (BME), is based on the end-to-end baseline model, E2E-Spot, as presented in~\cite{Hong2022Spotting}. We generate several variants of the E2E-Spot model to create a candidate model set and propose a strategy for selecting appropriate model members from this set while assigning appropriate weights to each selected model. More details can be found in~\cite{Wang2023ABoosted-arxiv}. The resulting ensemble model takes into account uncertainties in event length, optimal network architectures, and optimizers, making it more robust than the baseline model. Our approach has the potential to handle various video event analysis tasks.

\mysection{B3 - BASIK}\\
\textit{Juntae Kim, Gyusik Choi, Jeongae Lee, and Jongho Nang}\\
\textit{\{jtkim1211, gschoi, jalee3, jhnang\}@sogang.ac.kr}

Our primary focus is on tackling the class imbalance in the dataset of nine 90-minute soccer games, most frames of which are labelled as \textit{background}. We successfully implement Label Expansion to extend the labels of \textit{PASS} and \textit{DRIVE} frames to adjacent frames. The best performance is consistently achieved with a window size of four frames. The class imbalance is further mitigated through the application of the Focal Loss function, achieving optimal results with $\alpha$=1 and $\gamma$=2. Furthermore, we substitute the Gated Recurrent Unit in the original model with a Transformer encoder for better temporal reasoning. The final model is an ensemble of three models with different temporal reasoning architectures, contributing to a substantial 13.99\% improvement in test set performance compared to the baseline model.

\mysection{B4 - FC Pixel Nets}\\
\textit{Ikuma Uchida, 
Atom Scott, 
Taiga Someya, 
Kota Nakajima, 
Kenji Kobayashi, 
Hidenari Koguchi, 
and Ryuto Fukushima
}\\
\textit{uchida.ikuma@image.iit.tsukuba.ac.jp, 
\{atom.james.scott,atokota1022\}@gmail.com, 
\{taiga98-0809, kobayashi-kenji, hidenari-1108-hk, fukushima-ryuto0407\}@g.ecc.u-tokyo.ac.jp
}

In this Challenge, we aimed to enhance the E2E-Spot baseline method. We introduced RandomAffine, RandomPerspective and Mixup data augmentation during training, and adopted Focal Loss as the loss function. We conducted various experiments using the improved E2E-Spot model, including changes in input clip length and image size. We selected the top two models and applied Test-Time Augmentation to average their output class probabilities. Post-processing included the Savitzky-Golay filter, peak detection, and Non-Maximum Suppression. These improvements led to a significant increase in the mAP@1 metric score, reaching 83.53\%.

Additionally, we implemented a two-stage action recognition architecture using E2E-Spot models trained on optical flow, grayscale stacked frames, and RGB frames. We also tried combining YOLOv8 player detection with PySceneDetect for scene transition identification, to extract replay scenes by identifying camera zoom-ins based on a thresholded average ratio of the player's bounding box to image size. However, these attempts had limited impact on performance.

\subsection*{Dense Video Captioning} \label{app:densevideocaptioning}

\mysection{D3 - justplay}\\
\textit{Wei Dai,  Yongqiang Zhu, and Menglong Li}\\
\textit{loveispdvd@gmail.com, alexzhu.vip@gmail.com, mlli8803@163.com}

The baseline for this task is Temporally-Aware Feature Pooling for Dense Video Captioning in Video Broadcasts~\cite{Mkhallati2023SoccerNetCaption}. This approach divides the task of dense video captioning (DVC) into two stages: spotting and captioning. The first stage involves locating the events that need to caption, while the second stage involves generating captions for these events. Therefore, the baseline requires training two models: a spotting model and a captioning model. Pre-trained weights for both models are provided in the GitHub repository for this task.
To improve the performance of our captioning model, we experimented with replacing the pooling layer in the model. Additionally, we fine-tuned both the spotting and captioning models for a few epochs. The resulting METEOR score was 21.2, which was higher than the score of baseline1 but still lower than the score of baseline2.

\subsection*{Camera Calibration} \label{app:cameracalibration}

\mysection{C2 - Spiideo}\\
\textit{Håkan Ardö}\\
\textit{hakan.ardo@spiideo.com}

The camera parameters are estimated from the images in two steps. A pixel-level segmentation followed by a camera optimization using differential rendering. Code is available at \url{https://github.com/Spiideo/soccersegcal}.
The segmentation is performed using a DeepLabV3 CNN, segmenting the image into $6$ classes, representing different parts of the field. Sigmoid activations are used, which allows the different classes to overlap. Ground truth segmentations are generated from the SoccerNet annotations using floodfill operations.
A synthetic image segmentation is generated by rendering a 3D soccer-pitch and given camera parameters. The differential renderer SoftRasterizer\cite{Liu2019Soft} is used and the camera parameters updated using the local optimizer AdamW. To initiate the process a set of predefined start cameras are found by clustering all the cameras produced by the SoccerNet baseline method on the training data into 20 clusters using k-means. They are then pre-optimized using a loss that align the centers of the measured and rendered segments.

\mysection{C3 - SAIVA\_Calibration}\\
\textit{Hankyul Kim}\\
\textit{harry.kim@aibrain.co.kr}

SAIVA (Soccer AI Virtual Assistant) is an advanced AI platform designed for soccer match video analysis, with a key feature being the camera calibration module that supports other SAIVA modules with precise location data. This module offers two calibration methods: object detection, trained on a homography matrix with field transformation data, and keypoint detection, trained on $221$ annotated soccer field keypoints. These methodologies were synergized in the SoccerNet Camera Calibration 2023 Challenge. Here, the predicted homography matrix assists in choosing keypoints, evaluated based on distance from segmentation lines. The module scored $0.52$ and $0.53$ respectively in the test and challenge set evaluations, demonstrating its proficiency.

\mysection{C6 - NASK}\\
\textit{Kamil Synowiec}\\
\textit{kamil.synowiec@nask.pl}

The proposed approach consists of two main parts. Firstly, annotated points belonging to soccer pitch elements were transformed to segmentation masks by utilizing curve fitting techniques. These masks served as input for an instance segmentation model - Mask2Former with Swin-S hierarchical vision transformer as its backbone. Model was trained to detect and classify various field elements, including lines, conics, and goal parts. Subsequently, specific pitch points were localized by identifying the intersection of lines and ellipses derived from output segmentation masks. To compute the homography matrix, at least four image points mapped to corresponding points from the 3D pitch model are required. This matrix was estimated using RANSAC solver with different values of maximum reprojection threshold. Based on calculated homography, camera parameters were extracted and further refined using Perspective-n-Point (PnP) solver to obtain final results.

%\subsection*{Player Re-Identification} \label{app:reid}

\subsection*{Multiple Player Tracking} \label{app:tracking}

\mysection{T2 - Enhance End-to-End Multi-Object Tracking by CO-MOT}\\
\textit{Feng Yan, Weixin Luo, Yiyang Gan, Yujie Zhong, and Lin Ma}\\
\textit{\{yanfeng05, luoweixin, ganyiyang, zhongyujie, malin11\}@meituan.com}

We propose a effective approach to enhance end-to-end multi-object tracking based on motrv2~\cite{Zhang2022MOTRv2-arxiv}, by incorporating a coopetition label assignment proposed by CO-MOT~\cite{Yan2023Bridging-arxiv}. To address the imbalance between positive and negative samples for detection queries caused by Tracking Aware Label Assignment, especially in the closed environment of SoccerNet matches where all objects are detected in the initial frame and there are few new objects in subsequent frames, we introduce the Copetition Label Assignment. In the first five layers of the decoder, detection queries are responsible not only for detecting new objects but also for detecting already tracked objects. This significantly increases the number of positive samples and effectively trains the detection queries. Thanks to the self-attention mechanism in the decoder, the performance of the detection queries is transferred to the tracking queries, further improving tracking performance. We achieved a 69.5 HOTA on the Tracking 2023 challenge data without using any additional data at \href{https://github.com/BingfengYan/CO-MOT}{https://github.com/BingfengYan/CO-MOT}

\mysection{T3 - MOT4MOT~\cite{Shitrit2023SoccerNet-arxiv}}\\
\textit{Ishay Be'ery, Gal Shitrit, and Ido Yerushalmy}\\
\textit{\{ishaybee, galshi, idoy\}@amazon.com}

For player tracking, we employ a state-of-the-art online multi-object tracker DeepOCSORT along with a fine-tuned YOLO8 object detector. We finetuned an appearance model on a well curated dataset from the training set. To overcome the limitations of the online approach, we incorporate a post-processing stage that includes interpolation and appearance free track merging. Additionally, an appearance-based track merging technique is used to handle track termination and creation far from the image boundaries. For ball tracking, we treat it as a single object detection problem and utilize a fine-tuned YOLOv8l detector. For training the detector we curate the training data from erroneous labels using pre-trained ball detector and use only well annotated frames. In addition, we used filtering techniques such as estimating the ball trajectory as 3rd order polynomial with a large temporal window and rejecting detections that are far from this trajectory to enhance detection precision. More information is available in our technical report~\cite{Shitrit2023SoccerNet-arxiv}.

\mysection{T4 - ICOST}\\
\textit{Jiajun Fu, Jinghang Xu, Wending Zhao, Lizhi Wang, and Jianqin Yin}\\
\textit{\{JaakkoFu,xjh\_amber,windy,wanglizhi,jqyin\} @bupt.edu.cn}

The first step for our method is to detect balls, players, and referees in the image. Since the blurry image caused by camera or player movement will introduce missing detections, we first run a state-of-the-art image deblurring method to pre-preprocess images. Then, we trained a YOLOX detector~\cite{Ge2021YOLOX-arxiv} with the preprocessed training data. This can prevent false detections like the spectators and ball boys. The second step is to run a short-term tracking algorithm. We adopted OC-SORT~\cite{Cao2023Observation} and introduced a Buffered Complete Intersection of Union (BCIoU) for the association between detections and tracklets. We also integrate Camera Motion Compensation. Finally, we introduce Co-occurrence-aware Hierarchical Clustering to merge tracklets, where the mergence between two tracklets with co-occurrent detections is prohibited. The clustering is conducted by comparing the appearance features.  The appearance features are extracted from a reid network.

\mysection{T5 - SAIVA\_Tracking}\\
\textit{Byoungkwon Lim, Yeeun Joo, Seungcheon Lee}\\
\textit{bklim@aibrain.co.kr, yeeun.joo@turingai.global, sclee@turingai.global}

As part of SAIVA (Soccer AI Virtual Assistant), the player’s tracking system is based on ByteTrack and four added features - Camera movement estimation, IoU based distance, Order distance and Outside player calibration. The Camera movement estimation focuses on adjusting distances between tracks and detecting objects. This takes place by estimating and comparing paired objects on consecutive frames. The IoU based distance is developed for estimating distances between objects without overlapping. In that case, its distance value will be a constant which makes two areas close together by enlarging its area. The Order distance is being combining with IoU based distance and used to minimize object detection errors. It calculates distances, using each object’s X,Y axis order as elements. The Outside player calibration improves focus on the tracking objects by eliminating inactive tracks if the duration of outside detection is over a certain threshold. SAIVA tracking system records HOTA $63.19$ about SoccerNet Player Tracking Challenge 2023

\mysection{T6 - ZTrackers}\\
\textit{Ibrahim Salah, Mohamed Abdelwahed, Abdullah Kamal, Mohamed Nasr, and Amr Abdelaziz}\\
\textit{\{s-ibrahimsaad,s-mohamedabdelwahed,  \\s-abdullahkamal,s-mohamed\_nasr, \\s-amr.ragab1041\}@zewailcity.edu.eg}

Our methodology involved combining the YOLOv5 Medium object detection model with the ByteTrack algorithm. We trained the YOLOv5 Medium model on annotated soccer game images to accurately detect players and the ball. The detections obtained from YOLOv5 Medium were then passed to the ByteTrack algorithm. We fine-tuned several parameters, including the tracking confidence threshold, frames buffer, and matching threshold, to optimize tracking accuracy, smoothness, and robustness. Additionally, we fine-tuned the parameters of the Kalman filter to enhance velocity and position estimates. By integrating YOLOv5 Medium, ByteTrack, and the fine-tuned Kalman filter parameters, we developed a robust system capable of providing real-time and accurate player and ball tracking in the soccer video game

\subsection*{Jersey Number Recognition} \label{app:jerseynumber}

\mysection{J2 - CLIP Zero-Shot Jersey Number Labeling}\\
\textit{Konrad Habel, Fabian Deuser, and Norbert Oswald}\\
\textit{konrad.habel@unibw.de, fabian.deuser@unibw.de, norbert.oswald@unibw.de}

Our approach uses an ensemble of two CLIP~\cite{Radford2021Learning} pre-trained models fine-tuned on the SoccerNet Re-Identification dataset instead of the tracklets of the actual SoccerNet Jersey Number Recognition dataset. Due to the high amount of label noise in the dataset of the challenge, we decided to train our models not on the given data and labels. Instead, we use the ViT-L14 CLIP model of OpenAI for automatically zero-shot labeling the more diverse SoccerNet Re-Identification dataset without any human annotation. On these pseudo labels per image we fine-tuned the image encoders of the two models. To predict the jersey number on a tracklet basis, we use a majority voting taking only images with a classification probability over $70\%$ for numbers $1$~-~$99$ into account. We achieve on the Test set an accuracy of $90.09\%$ and on the Challenge set $90.95\%$.

\mysection{J3 - zzzzz}\\
\textit{Junjie Li, Guanshuo Wang, Fufu Yu, Qiong Jia, and Shouhong Ding}\\
\textit{serenitycapo@gmail.com, \{mediswang, fufuyu, boajia, ericshding\}@tencent.com}

We approached this challenge by employing various problem formulation methodologies, which involved formulating the task as an optical character recognition (OCR) problem and modeling the recognition of jersey numbers as a sequential prediction based on tracklets. To begin with, we utilized a state-of-the-art text detection model called Deepsolo to obtain initial OCR results. Building upon these predictions at the image level, the sequential prediction model has been designed in a similar manner to transformer, where the image patches are replaced by feature representations of images in a tracklet. To further boost the performance, such as performing model ensemble with multi-input resolution. Finally, the sequential prediction results are refined with the original text detection results to obtain the final predictions.

\mysection{J5 - MT-IOT}\\
\textit{Gan Yiyang, Luo Weixin, Yan Feng, Lin Ma}\\
\textit{\{ganyiyang, luoweixin, yanfeng05\}@meituan.com, forest.linma@gmail.com}

The accurate recognition of jersey numbers in soccer broadcasts is important for tracking player movements, evaluating performance, and making strategic decisions during games. To address this task, the authors propose a multi-task video classification approach that leverages temporal and spatial cues from player tracklets. They employ an advanced video transformer network as a powerful backbone to extract features from tracklets and designed a multi-task classification head to address the issue of long-tailed data distribution. The task is divided into two sub-tasks: predicting the two digits of the number separately and predicting the permutation of the digits. The authors use binary cross-entropy loss for the digit head and cross-entropy loss for the permutation head to effectively address the issue of unbalanced data. The proposed approach achieves competitive accuracy of 87.37\% on the test set and 81.70\% on the challenge set, showcasing its effectiveness.

\mysection{J6 - justplay}\\
\textit{Wei Dai, Yongqiang Zhu, and Menglong Li}\\
\textit{loveispdvd@gmail.com, alexzhu.vip@gmail.com, mlli8803@163.com}

The first step is to detect the jersey numbers present in the image. This is a necessary step before proceeding with recognition. In the second step, we manually reviewed over $700$ folders in both the training and testing sets, which contained the cropped jersey number patches during the detection phase. Each folder only retained the patches with the same number as the folder's annotation, while removing those with different detected numbers and those with irrelevant contents due to false detection. For the folder labeled with no jersey number, we added some patches without jersey numbers. This allows the model to recognize results for no jersey number as well. Finally, we inferred on the challenge set using the fine-tuned detection model and the fine-tuned recognition model. The majority of recognition results in each folder were taken as the predicted result for that folder. After submission, the accuracy score was $77.77$.

\mysection{J7 - AIBrain Global Team}\\
\textit{Iftikar Muhammad and Hasby Fahrudin}\\
\textit{\{iftikarm,hfahrudin\}@aibrain.co.kr}

As an effort of SAIVA (Soccer AI Virtual Assistant), we propose an approach to accurately detect and recognize jersey numbers by leveraging player body orientation information. Our method aims to emulate human perception when observing tracklet images to determine the jersey numbers. Instead of relying on a large number of tracklet images, we utilize a confidence sorting algorithm based on the visibility of the jersey number and the quality of the image.
First to tackle the low-resolution issue we upscale the image using ESRGAN-based model, and then we utilize keypoints extracted from pose-estimation model to localize the jersey number and get the body orientation. Lastly we use body orientation and image quality assessment to rank the prediction confidence of each tracklet. On SoccerNet Challenge 2023 our work achieved 76.05\% on Test-set and 75.17\% on Challenge-set in terms of classification accuracy.


\end{document}